\documentclass{article}
\usepackage{spconf,amsmath,graphicx,hyperref}
\usepackage{booktabs, colortbl, multirow, arydshln}
\usepackage{tcolorbox}

\definecolor{skyblue}{RGB}{203, 221, 245}
\newcommand{\skyblue}{\rowcolor{skyblue}}

\title{SAMAS: A Spectrum-Guided Multi-Agent System for Achieving Style Fidelity in Literary Translation}
\name{Jingzhuo Wu$^{1,*}$\thanks{*Equal contribution.}, Jiajun Zhang$^{2,*}$, Keyan Jin$^{3}$, Dehua Ma$^{4}$, Junbo Wang$^{5}$}

\address{
  $^{1}$Beijing Normal University \qquad
  $^{2}$University of Science and Technology of China \\
  $^{3}$University of Coimbra \qquad
  $^{4}$Beijing University of Posts and Telecommunications \\
  $^{5}$Northwestern Polytechnical University
}
\usepackage{xcolor}       
\usepackage{tcolorbox}    

\begin{document}

\newtcbox{\roundbox}{on line, 
    colback=gray!10,      
    colframe=gray!50,     
    boxrule=0.5pt,        
    arc=2pt,              
    outer arc=2pt,        
    boxsep=1pt,           
    left=1pt, right=1pt, top=0.5pt, bottom=0.5pt 
}
\maketitle
\begin{abstract}
Modern large language models (LLMs) excel at generating fluent and faithful translations. However, they struggle to preserve an author's unique literary style, often producing semantically correct but generic outputs. This limitation stems from the inability of current single-model and static multi-agent systems to perceive and adapt to stylistic variations. To address this, we introduce the \textbf{\underline{S}}tyle-\textbf{\underline{A}}daptive \textbf{\underline{M}}ulti-\textbf{\underline{A}}gent \textbf{\underline{S}}ystem (SAMAS), a novel framework that treats style preservation as a signal processing task. Specifically, our method quantifies literary style into a Stylistic Feature Spectrum (SFS) using the wavelet packet transform. This SFS serves as a control signal to dynamically assemble a tailored workflow of specialized translation agents based on the source text's structural patterns. Extensive experiments on translation benchmarks show that SAMAS achieves competitive semantic accuracy against strong baselines, primarily by leveraging its statistically significant advantage in style fidelity. 
\end{abstract}
\section{Introduction}
\label{sec:intro}


Literary translation remains a significant challenge in machine translation, often considered the field's ``\textit{final frontier}" as it requires a deep understanding of cultural context, rhetorical figures, and unique authorial styles~\cite{transagent}. To properly evaluate translation quality in this complex domain, it is helpful to use the three traditional principles of translation. These principles provide a clear framework, beginning with \textbf{Fidelity}, which refers to the accurate transfer of the source text's core meaning. Next, \textbf{Fluency} ensures that the translated text reads naturally and smoothly in the target language. Finally, \textbf{Felicity} involves the more nuanced task of capturing the original's style, tone, and literary grace.

While advancing LLMs~\cite{gpt4,gemini1.5,zhang2025plotcraft,zheng2026should} and translation agents~\cite{zhang2025tactic,xuan2025translaw,wang2025evoc2rust} show great promise, with models generating highly fluent text, frameworks like TransAgents improving quality via collaboration~\cite{transagent}, and stylometry tools adeptly identifying authorial markers~\cite{briakou2024stepbystep, zhang2024evolving}, they excel at Fidelity and Fluency. However, they fail to achieve \textbf{Felicity}, as they are constrained by paradigms that fundamentally struggle with literary style. For instance, these Multi-Agent Systems often use static, role-based workflows, while computational stylometry features remain descriptive rather than serving as dynamic controls. This systemic inflexibility leads to translations that, while semantically correct, are stylistically bland and fail to capture the author's unique ``\textit{fingerprint}" of rhythm, prosody, and structural complexity.

To fill this gap, we introduce the Style-Adaptive Multi-Agent System (SAMAS), an approach that reframes stylistic translation as a signal processing and control problem. Our method is built on two core innovations:
\textbf{(1) The Stylistic Feature Spectrum (SFS)}. We propose the SFS, a multi-scale feature vector extracted from a text's word-length sequence using the Wavelet Packet Transform (WPT). As a tool adept at analyzing non-stationary, multi-scale signals, WPT effectively captures the dynamic rhythmic properties of literary texts. This process transforms the abstract concept of ``style" into a computable and interpretable control signal.
\textbf{(2) An SFS-Driven Dynamic Architecture}. We design a multi-agent system where the SFS signal guides a deterministic routing mechanism to dynamically construct an optimal processing pipeline for each text segment. This approach fundamentally shifts from the static collaboration models of previous multi-agent systems to a real-time, adaptive workflow allocated based on the intrinsic style of the input data. Our extensive experiments on standard translation benchmarks confirm the effectiveness of our approach. The results demonstrate that SAMAS achieves a significant improvement in translation accuracy over strong baselines. This advantage is achieved without compromising semantic quality, which remains competitive with SOTA models. Rigorous human evaluations further corroborate these findings, confirming the superiority of the translations produced by SAMAS.


Our main contributions are as follows:
\vspace{-3pt}
\begin{itemize}  
    \item We are the first to quantify literary style as a signal to route multi-agent translation.
    \vspace{-1em}
    \item We design a dynamic multi-agent system (SAMAS) that uses the Wavelet Packet Transform (WPT) to generate a Stylistic Feature Spectrum (SFS). This signal dynamically configures a custom translation workflow, marking a shift from static to style-adaptive paradigms.
    \vspace{-1.5em}
    \item We conducted comprehensive experiments that demonstrate SAMAS significantly outperforms strong baselines in translation accuracy, a result confirmed by rigorous automatic metrics and human evaluations.
\end{itemize}

\begin{figure*}
\label{fig:main}
\centering
    \includegraphics[width=0.90\linewidth]{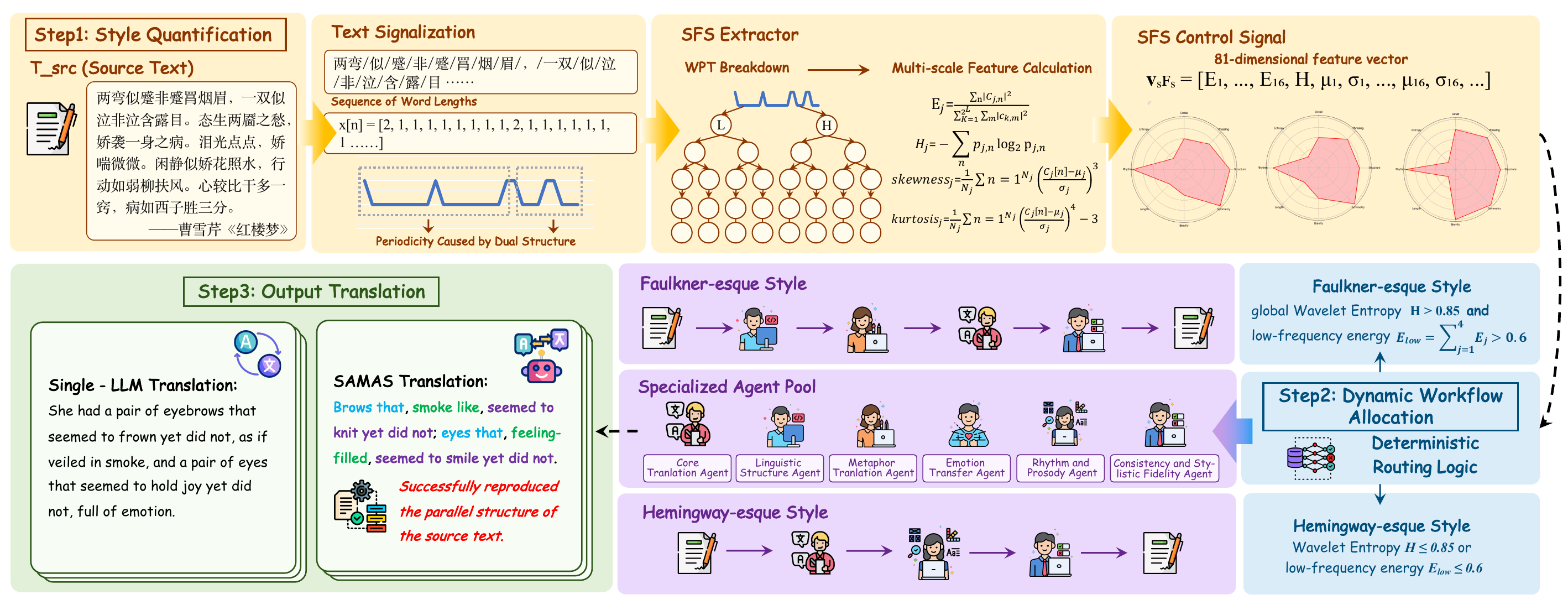}
    \caption{Overview of the SAMAS. The system first quantifies a text's literary style into a Stylistic Feature Spectrum (SFS) signal. This signal then guides the dynamic allocation of a workflow, which is executed by a pool of specialized agents to produce a stylistically faithful translation that outperforms a single-LLM baseline.}
    \vspace{-1.5em}
\end{figure*}

\section{Methodology}
\label{sec:format}

To address the challenge of inadequate stylistic fidelity in literary translation, we propose the Style-Adaptive Multi-Agent System (SAMAS). Our system consists of two core components: a Stylistic Feature Spectrum (SFS) that converts literary style into a computable signal, and a dynamic multi-agent framework that uses this signal to guide text generation. We detail the SFS in Section \ref{sec:sfs} and the multi-agent framework in Section \ref{sec:mas}. The overall architecture is shown in Figure \ref{fig:main}.

\subsection{Stylistic Feature Spectrum} 
\label{sec:sfs}
To treat literary style as a signal, we first convert a text segment into a numerical sequence based on word length. This sequence effectively captures the text's rhythm and structural complexity~\cite{schmidt2025teambettercallclaudestylechange, zhang2026completion}. Our choice of word length is motivated by its established role as a robust, language-agnostic proxy for stylistic rhythm in computational stylometry. This feature directly reflects an author's pacing and structural complexity, making it ideal for our signal-based analysis~\cite{pimentel2023revisitingoptimalitywordlengths}.  Since literary style is non-stationary, we use the Wavelet Packet Transform (WPT), which provides a multi-resolution analysis ideal for identifying stylistic patterns at various scales.

From the WPT coefficients, we construct a compact feature vector, the \textbf{Stylistic Feature Spectrum (SFS)} ($\mathbf{v}_{\text{sfs}}$), by extracting key features from each of the $2^L$ sub-bands (where $L$ is the decomposition level). The primary features include:
\begin{itemize}
    \item \textbf{Relative Wavelet Energy (RWE)}, as shown in equation \ref{equa1}, which measures the energy proportion in each sub-band $j$, reflecting the dominance of patterns at a specific scale.
    \begin{equation}
    \label{equa1}
        E_j = \frac{\sum_{n} |c_{j,n}|^2}{\sum_{k=1}^{2^L} \sum_{m} |c_{k,m}|^2}
    \end{equation}
    where $c_{j,n}$ is the $n$-th coefficient in the $j$-th sub-band.
    
    \item \textbf{Wavelet Entropy (WE)}, as shown in equation \ref{equa2}, which quantifies the disorder of the signal's energy distribution, indicating textual complexity.
    \begin{equation}
    \label{equa2}
        H_j = - \sum_{n} p_{j,n} \log_2 p_{j,n}
    \end{equation}
    where $p_{j,n}$ is the energy probability distribution of the coefficients in sub-band $j$.
    
    \item \textbf{Statistical Moments.} We also compute the mean, standard deviation, skewness (equation \ref{equa3}) and kurtosis (equation \ref{equa4}) of the coefficients in each sub-band to capture finer textural characteristics.
     \begin{equation}
     \label{equa3}
     skewness_j = \frac{1}{N_j} \sum{n=1}^{N_j} \left(\frac{c_j[n] - \mu_j}{\sigma_j}\right)^3 \end{equation}      
     \begin{equation}
     \label{equa4}
     kurtosis_j = \frac{1}{N_j} \sum{n=1}^{N_j} \left(\frac{c_j[n] - \mu_j}{\sigma_j}\right)^4 - 3 
     \end{equation}  

\end{itemize}
For a decomposition level of $L=4$, we extract features from all 16 sub-bands. Specifically, we compute 5 statistical features for each sub-band, consisting of Relative Wavelet Energy, mean, standard deviation, skewness, and kurtosis. We calculate one global Wavelet Entropy. The final SFS vector ($\mathbf{v}_{\text{sfs}}$) is therefore an 81-dimensional vector that forms a comprehensive, quantitative stylistic signature. This signature is a deterministic control signal for our translation system.

\subsection{SFS-Driven Dynamic Multi-Agent System}
\label{sec:mas}

Our SFS vector guides a dynamic multi-agent system to construct an optimal translation workflow for each text segment, enabling style adaptation. The system is composed of a routing mechanism and a pool of specialized agents.

\subsubsection{System Architecture}
The core of our system is a deterministic routing mechanism. It uses the input text's SFS vector ($\mathbf{V}_{\text{sfs}}$) to select an appropriate processing workflow from a predefined library. This rule-based approach, similar to a decision tree, maps quantitative stylistic features to specific translation strategies, making the system's logic inherently interpretable.

The workflows are constructed from a pool of six specialized agents, each designed to handle a distinct aspect of the translation. The \roundbox{\texttt{Core Translation Agent}} is responsible for fundamental semantic conversion, while the \roundbox{\texttt{Linguistic Structure Agent}} manages complex syntax. Stylistic nuances are addressed by the \roundbox{\texttt{Metaphor}} \roundbox{\texttt{Translation Agent}} for rhetorical figures, the \roundbox{\texttt{Emotion}} \roundbox{\texttt{Transfer Agent}} for preserving tone, and the \roundbox{\texttt{Rhythm}} \roundbox{\texttt{and Prosody Agent}} for reproducing the text's cadence. Finally, a \roundbox{\texttt{Consistency and Stylistic}} \roundbox{\texttt{Fidelity Agent}} ensures overall stylistic and terminological coherence. Each agent is an instance of the respective base model, guided by a role-specific prompt.

\subsubsection{Dynamic Workflow Allocation}
The system's adaptiveness is best illustrated by how it handles texts with distinct styles.

\textbf{Faulkner-esque Style}, characterized by high complexity and macro-structure dominance, corresponds to an SFS vector with a Wavelet Entropy $H>0.85$ and cumulative low-frequency energy $E_{\text{low}} = \sum_{j=1}^{4} E_{j} > 0.6$.

These thresholds were determined on a 200-segment validation set from two authors with contrasting styles (William Faulkner and Ernest Hemingway) to enable a robust, controlled analysis. For each segment, we computed its SFS vector. A grid search and ROC analysis identified the optimal thresholds for Wavelet Entropy (H) and cumulative low-frequency energy ($\mathbf{E}_{\text{low}}$) that yielded the highest classification accuracy on this set. The selected values, $H>0.85$ and $E_{low}>0.6$, represent the empirical points that best separate the two stylistic classes on the validation set.

Grounded in this empirical data, these thresholds serve to confirm high complexity and macro-structure dominance in a text. The routing mechanism identifies this signature and deploys a specialized workflow, such as an ordered sequence of agents: \textit{Linguistic Structure}, \textit{Metaphor Translation}, \textit{Core Translation}, and \textit{Consistency Check}. This strategy deconstructs stylistic complexity before the main translation task.

\textbf{Hemingway-esque Style}, characterized by structural simplicity and a short, rhythmic pace, corresponds to an SFS vector with low Wavelet Entropy and balanced high-frequency energy. A text is classified as Hemingway-esque if it does not meet the criteria for the Faulkner-esque style. This means its Wavelet Entropy $H\leq0.85$ or its low-frequency energy $E_{low}\leq0.6$. The system then selects a more direct workflow, such as: \textit{Core Translation}, \textit{Rhythm and Prosody}, and \textit{Consistency Check}. This efficient pipeline bypasses unneeded steps to focus resources on preserving the key feature of concise rhythm.

Our SFS-driven system uses the input's stylistic signature to deliver adaptive translation, moving beyond static models.

\vspace{-1em}
\section{Experiments}
\label{sec:exp}

\begin{table*}[h]
\centering
\scriptsize 
\setlength{\tabcolsep}{10.5pt} 
\vspace{-1em}

\begin{tabular}{@{}lccccccccc@{}}
\toprule
\multirow{3}{*}{\textbf{Models}} & \multirow{3}{*}{\textbf{Routing Method}} & \multicolumn{4}{c}{\textbf{FLORES-200}} & \multicolumn{4}{c}{\textbf{WMT24}} \\
\cmidrule(lr){3-6} \cmidrule(lr){7-10}
& & \multicolumn{2}{c}{\textbf{en$\rightarrow$xx}} & \multicolumn{2}{c}{\textbf{xx$\rightarrow$en}} & \multicolumn{2}{c}{\textbf{en$\rightarrow$xx}} & \multicolumn{2}{c}{\textbf{xx$\rightarrow$en}} \\
\cmidrule(lr){3-4} \cmidrule(lr){5-6} \cmidrule(lr){7-8} \cmidrule(lr){9-10}
& & \textbf{XC} & \textbf{KW} & \textbf{XC} & \textbf{KW} & \textbf{XC} & \textbf{KW} & \textbf{XC} & \textbf{KW} \\
\midrule
\skyblue \multicolumn{10}{c}{\textit{\textbf{Single-LLM}}} \\
\midrule  
Qwen3-235B-A22B~\cite{qwen3} & - & 84.17 & 78.25 & 94.05 & 87.17 & 73.97 & 68.57 & 82.59 & 77.83 \\
Qwen3-30B-A3B~\cite{qwen3} & - & 77.53 & 70.18 & 91.55 & 85.12 & 66.97 & 59.97 & 81.21 & 77.11 \\
Qwen-mt-plu~\cite{qwen3} & - & 86.97 & 81.27 & 94.43 & 87.43 & 76.21 & 70.87 & 84.37 & 79.40 \\
GPT-5~\cite{gpt4} & - & 95.46 & 90.83 & \underline{96.70} & 88.61 & 86.83 & 81.03 & 87.73 & 82.33 \\
GPT-4.1~\cite{gpt4} & - & 91.45 & 86.14 & 95.71 & 88.20 & 81.23 & 75.81 & 86.00 & 80.48 \\
Gemini-2.5-pro~\cite{gemini1.5} & - & 90.87 & 86.33 & 95.02 & 87.82 & 79.92 & 75.42 & 83.59 & 78.74 \\
DeepSeek-V3~\cite{deepseek-v3} & - & 84.43 & 89.80 & 95.14 & 88.45 & 84.71 & 79.61 & 83.20 & 80.99 \\
DeepSeek-R1~\cite{deepseek-v3} & - & 89.33 & 90.75 & 95.13 & 88.36 & 86.10 & 80.45 & 83.98 & 75.01 \\
\midrule
\skyblue \multicolumn{10}{c}{\textit{\textbf{Multi-Agent Frameworks}}} \\
\midrule
HiMATE \cite{HiMATE}& - & 77.50 & 69.73 & 91.06 & 85.00 & 65.25 & 69.03 & 78.44 & 75.09 \\
MAATS \cite{MAATS}& - & 85.44 & 79.63 & 94.09 & 87.21 & 75.03 & 69.62 & 82.66 & 77.93 \\
TransAgents \cite{transagent} & - & 87.65 & 81.87 & 94.31 & 87.28 & 76.61 & 71.22 & 84.36 & 79.48 \\
TACTIC \cite{tatic} & - & 96.19 & \underline{92.64} & 96.69 & 90.15 & 86.95 & 81.26 & 89.07 & 82.46 \\
\midrule
\skyblue \multicolumn{10}{c}{\textit{\textbf{SAMAS}}} \\
\midrule
Qwen3-235B-A22B~\cite{qwen3} & SAMAS & \underline{96.78}\textsuperscript{*} & \textbf{92.97}\textsuperscript{*} & 95.86 & \textbf{90.84}\textsuperscript{*} & \underline{87.10}\textsuperscript{*} & \underline{83.15}\textsuperscript{*} & 87.13 & 82.07 \\
Qwen3-235B-A22B~\cite{qwen3} & Faulkner-esque Style & 91.33 & 86.68 & 95.22 & 89.39 & 81.12 & 76.25 & 86.10 & 80.74 \\
Qwen3-235B-A22B~\cite{qwen3} & Hemingway-esque Style & 95.86 & 89.17 & 94.01 & 89.80 & 82.97 & 78.06 & 86.66 & 81.56 \\
Qwen3-30B-A3B~\cite{qwen3} & SAMAS & 93.48 & 88.72 & 95.96 & 90.15 & 86.85 & 81.26 & 89.07 & 82.06 \\
Qwen3-30B-A3B~\cite{qwen3} & Faulkner-esque Style & 92.19 & 88.64 & 95.82 & 88.26 & 82.71 & 78.08 & 86.17 & 80.35 \\
Qwen3-30B-A3B~\cite{qwen3} & Hemingway-esque Style & 91.56 & 88.22 & 95.37 & 88.00 & 81.68 & 77.38 & 85.85 & 80.20 \\
Qwen-mt-plu~\cite{qwen3} & SAMAS & 94.90 & 89.97 & 95.70 & 88.19 & 83.47 & 78.65 & \underline{89.15} & \underline{83.04} \\
Qwen-mt-plu~\cite{qwen3} & Faulkner-esque Style & 93.11 & 88.69 & 95.65 & 88.17 & 82.69 & 77.83 & 86.34 & 80.54 \\
Qwen-mt-plu~\cite{qwen3} & Hemingway-esque Style & 94.38 & 88.51 & 95.12 & 88.18 & 82.95 & 78.34 & 86.36 & 80.52 \\
GPT-5\cite{gpt4} & SAMAS & \textbf{96.93}\textsuperscript{*} & 91.72 &\textbf{97.10}\textsuperscript{*} & \underline{90.43}\textsuperscript{*} & \textbf{87.60}\textsuperscript{*} & \textbf{83.78}\textsuperscript{*} & \textbf{89.97}\textsuperscript{*} & \textbf{83.23}\textsuperscript{*} \\
GPT-5~\cite{gpt4} & Faulkner-esque Style & 95.08 & 90.32 & 96.06 & 88.08 & 84.20 & 82.48 & 87.62 & 81.68 \\
GPT-5~\cite{gpt4}& Hemingway-esque Style & 94.37 & 89.32 & 93.96 & 89.23 & 86.17 & 81.28 & 87.62 & 82.34 \\
\bottomrule
\end{tabular}
\caption{
    Comparison of translation performance on the FLORES-200 and WMT24 benchmarks, evaluated with XCOMET (XC) and COMETKIWI-23 (KW) metrics. The ``Models" column specifies the base LLM used for translation, while the ``Routing Method" column details the workflow applied within our SAMAS framework (e.g., dynamic routing or fixed style-based routes). ``Qwen3-235B-A22B" denotes the Instruct version. The \textbf{best} and \underline{second-best} scores are highlighted. Scores marked with * indicate a statistically significant improvement over the strongest multi-agent baseline (TACTIC) with p $<$ 0.05.
}
\label{tab:main_results}
\end{table*}

We evaluated the SAMAS framework on standard translation benchmarks and a curated, style-differentiated corpus.
\vspace{-1em}
\subsection{Experiments Setup}
 
\textbf{Baselines.} We compare SAMAS against two primary categories of baselines: single large language models (LLMs) and existing multi-agent translation systems.

\textbf{Datasets.} We evaluate our framework on two standard benchmarks: FLORES-200~\cite{costa2022no} and WMT24~\cite{kocmi-etal-2024-findings}. Our assessment covers English-centric translation (en$\leftrightarrow$xx) across five diverse language pairs: German (de), Japanese (ja), Russian (ru), Ukrainian (uk), and Chinese (zh). For WMT24, which only provides en$\rightarrow$xx data, we generated the reverse translations to ensure a comprehensive evaluation. All datasets are standardized using the Tower framework~\cite{alves2024tower}.

\textbf{Evaluation Metrics.} Our evaluation relies on three primary metrics with high correlation to human judgment: the reference-based XCOMET-XXL~\cite{guerreiro2024xcomet} and MetricX-24-XXL~\cite{juraska2024metricx24googlesubmissionwmt}, and the reference-free COMETKIWI-23-XXL~\cite{rei2023scaling}. We select the XXL variants of XCOMET and MetricX as they were used in the WMT24 official evaluation and offer better alignment with human scores. For a comprehensive comparison, we also report ChrF~\cite{popovic2015chrf} and sacreBLEU~\cite{post2018call} scores.

\vspace{-1em}
\subsection{Main Results}
As shown in Table \ref{tab:main_results}, SAMAS significantly outperforms strong baselines on FLORES-200 and WMT24. It substantially boosts base models, lifting the Qwen3-235B-A22B~\cite{hui2024qwen2} XCOMET score on FLORES-200 en$\rightarrow$xx from 84.17 to 96.78, surpassing the top multi-agent system TACTIC at 96.19 and powerful LLMs like GPT-5 at 95.46. Paired with GPT-5, SAMAS establishes new SOTA results with statistically significant gains over the strongest multi-agent baseline. An ablation study confirms dynamic routing outperforms fixed workflows. For instance, on Qwen3-235B-A22B, SAMAS scored 96.78, surpassing the fixed Faulkner-esque and Hemingway-esque workflows at 91.33 and 95.86, respectively. This result shows that adapting to the input signal is crucial, as no single strategy is universally optimal.

\begin{figure}
    \includegraphics[width=1.0\columnwidth]{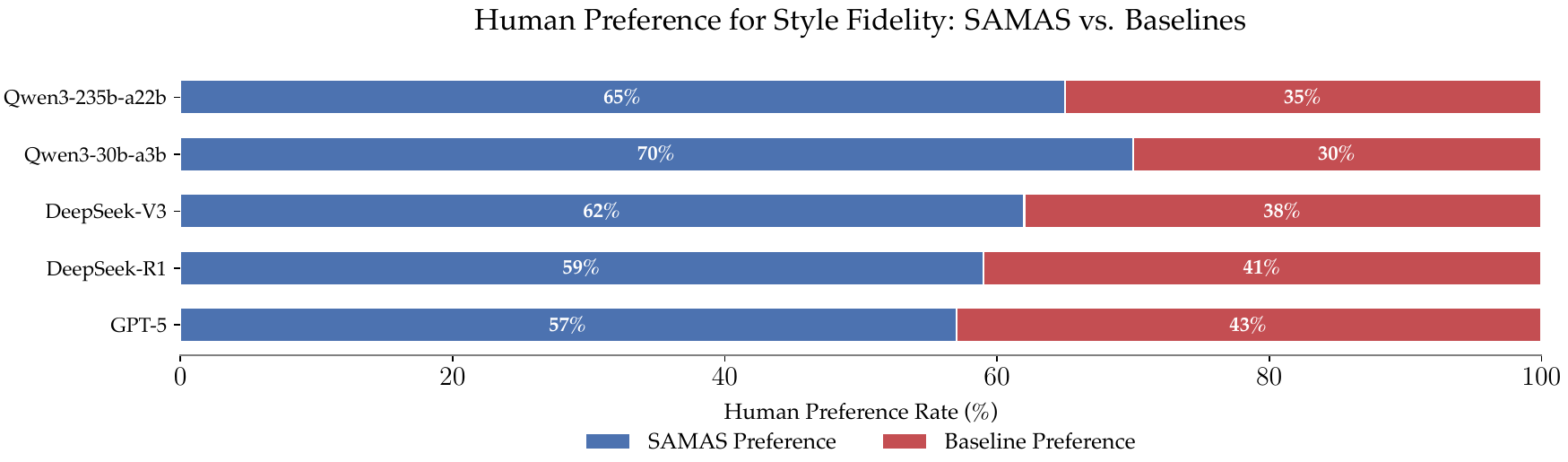}
    \caption{Human preference for style fidelity in head-to-head comparisons. In all evaluations, translations generated by SAMAS were consistently favored by human evaluators over those from five different strong baseline models.}
    \label{analysis}
    \vspace{-1em}
\end{figure}
\vspace{-2em}
\subsection{Style Analysis}
As shown in Figure \ref{analysis}, human evaluations confirm the stylistic superiority of SAMAS. In head-to-head comparisons for style fidelity, our system was consistently preferred by evaluators over all five baselines, including top-tier models like GPT-5. This result validates that our style-adaptive, multi-agent approach more effectively captures stylistic nuances than standard single-LLM methods.



\section{Conclusion}
\label{sec:conclusion}

In this work, we addressed the challenge of stylistic felicity in literary translation, a domain where current methods often fail despite achieving high semantic accuracy. We introduced the Style-Adaptive Multi-Agent System (SAMAS), which treats stylistic translation as a signal processing problem. Our approach quantifies literary style into a Stylistic Feature Spectrum (SFS) signal, which then dynamically configures a custom workflow executed by specialized agents.
Our experiments show that SAMAS significantly outperforms strong single-LLM and multi-agent baselines on standard translation benchmarks. Furthermore, human evaluations confirm that our system produces translations with superior style fidelity, with evaluators consistently preferring SAMAS in head-to-head comparisons. By modeling style as a dynamic control signal, SAMAS presents a promising shift from static to style-adaptive paradigms for complex text generation tasks.


\vfill\pagebreak

\bibliographystyle{IEEEbib}
\bibliography{strings,refs}

\end{document}